\documentclass{article}

\usepackage{booktabs}

\def\1{\bm{1}}

\def\rvg{{\mathbf{g}}}
\def\rvh{{\mathbf{h}}}

\DeclareMathAlphabet{\mathsfit}{\encodingdefault}{\sfdefault}{m}{sl}
\SetMathAlphabet{\mathsfit}{bold}{\encodingdefault}{\sfdefault}{bx}{n}

\def\gL{{\mathcal{L}}}

\def\gR{{\mathcal{R}}}

\usepackage{microtype}
\usepackage{graphicx}
\usepackage{enumitem}
\usepackage{booktabs} 

\usepackage{tikz}
\usepackage{subcaption}

\usepackage{pgfplots}

\usepackage{caption,subcaption}
\usepackage{mathtools}
\usepackage{amsthm}
\usepackage{multirow}
\usepackage{booktabs} 

\usepackage{hyperref}

\usepackage[accepted]{icml2024}

\usepackage{amsmath}
\usepackage{amssymb}
\usepackage{mathtools}
\usepackage{amsthm}

\usepackage[capitalize,noabbrev]{cleveref}

\newtheorem{remark}{Remark}
\newtheorem{hypothesis}{Hypothesis}

\usepackage[textsize=tiny]{todonotes}

\icmltitlerunning{MMPareto: Boosting Multimodal Learning with Innocent Unimodal Assistance}

\begin{document}

\twocolumn[
\icmltitle{MMPareto: Boosting Multimodal Learning with Innocent Unimodal Assistance}




\begin{icmlauthorlist}
\icmlauthor{Yake Wei}{yyy}
\icmlauthor{Di Hu}{yyy,comp}
\end{icmlauthorlist}

\icmlaffiliation{yyy}{Gaoling School of Artificial Intelligence, Renmin University of China, Beijing, China}
\icmlaffiliation{comp}{Beijing Key Laboratory of Big Data Management and Analysis Methods, Beijing, China}

\icmlcorrespondingauthor{Di Hu}{dihu@ruc.edu.cn}

\icmlkeywords{Machine Learning, ICML}

\vskip 0.3in
]



\printAffiliationsAndNotice{}  

\begin{abstract}
Multimodal learning methods with targeted unimodal learning objectives have exhibited their superior efficacy in alleviating the imbalanced multimodal learning problem. However, in this paper, we identify the previously ignored gradient conflict between multimodal and unimodal learning objectives, potentially misleading the unimodal encoder optimization. To well diminish these conflicts, we observe the discrepancy between multimodal loss and unimodal loss, where both gradient magnitude and covariance of the easier-to-learn multimodal loss are smaller than the unimodal one. With this property, we analyze Pareto integration under our multimodal scenario and propose MMPareto algorithm, which could ensure a final gradient with direction that is common to all learning objectives and enhanced magnitude to improve generalization, providing innocent unimodal assistance. Finally, experiments across multiple types of modalities and frameworks with dense cross-modal interaction indicate our superior and extendable method performance. Our method is also expected to facilitate multi-task cases with a clear discrepancy in task difficulty, demonstrating its ideal scalability. The source code and dataset are available at \url{https://github.com/GeWu-Lab/MMPareto_ICML2024}.
\end{abstract}

\section{Introduction}

People are immersed in a variety of sensors, encompassing sight, sound, and touch, which has sparked multimodal learning~\citep{baltruvsaitis2018multimodal,lin2023federated}, and audio-visual learning~\citep{wei2022learning}. Although many multimodal methods have revealed effectiveness, \citet{peng2022balanced} pointed out the imbalanced multimodal learning problem, where most multimodal models cannot jointly utilize all modalities well and the utilization of each modality is imbalanced. This problem has raised widely attention recently~\citep{zhang2024multimodal}. Several methods have been proposed to improve the training of worse learnt modality with additional module~\citep{wang2020makes} or modality-specific training strategy~\citep{peng2022balanced,wu2022characterizing,wei2024enhancing}. These methods often have one common sense that targetedly improves unimodal training. Among them, multitask-like methods that directly add unimodal learning objectives besides the multimodal joint learning objective, exhibit their superior effectiveness for alleviating this imbalanced multimodal learning problem~\citep{wang2020makes,du2023uni,fan2023pmr}.

\begin{figure*}[t]
\centering
	\begin{subfigure}[t]{.23\textwidth}
			\centering
			\includegraphics[width=\textwidth]{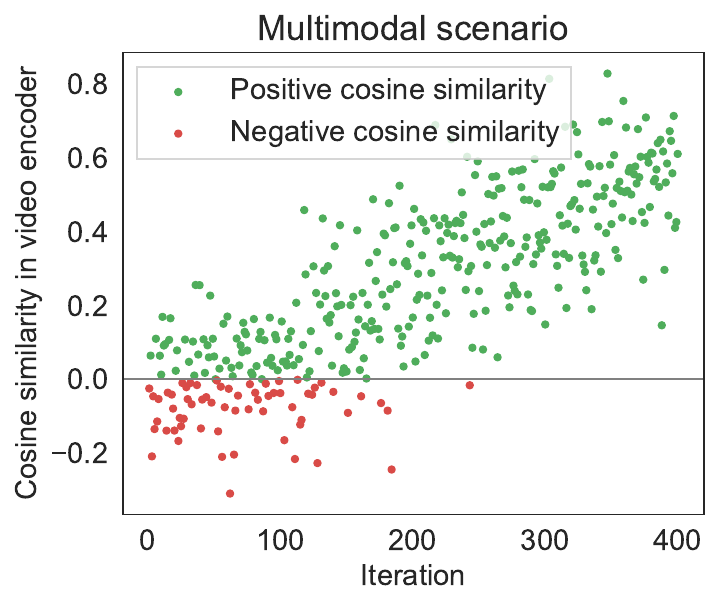}
			\caption{Kinetics Sounds.}
			\label{fig:teaser-conflic}
	\end{subfigure}
    \begin{subfigure}[t]{.23\textwidth}
			\centering
			\includegraphics[width=\textwidth]{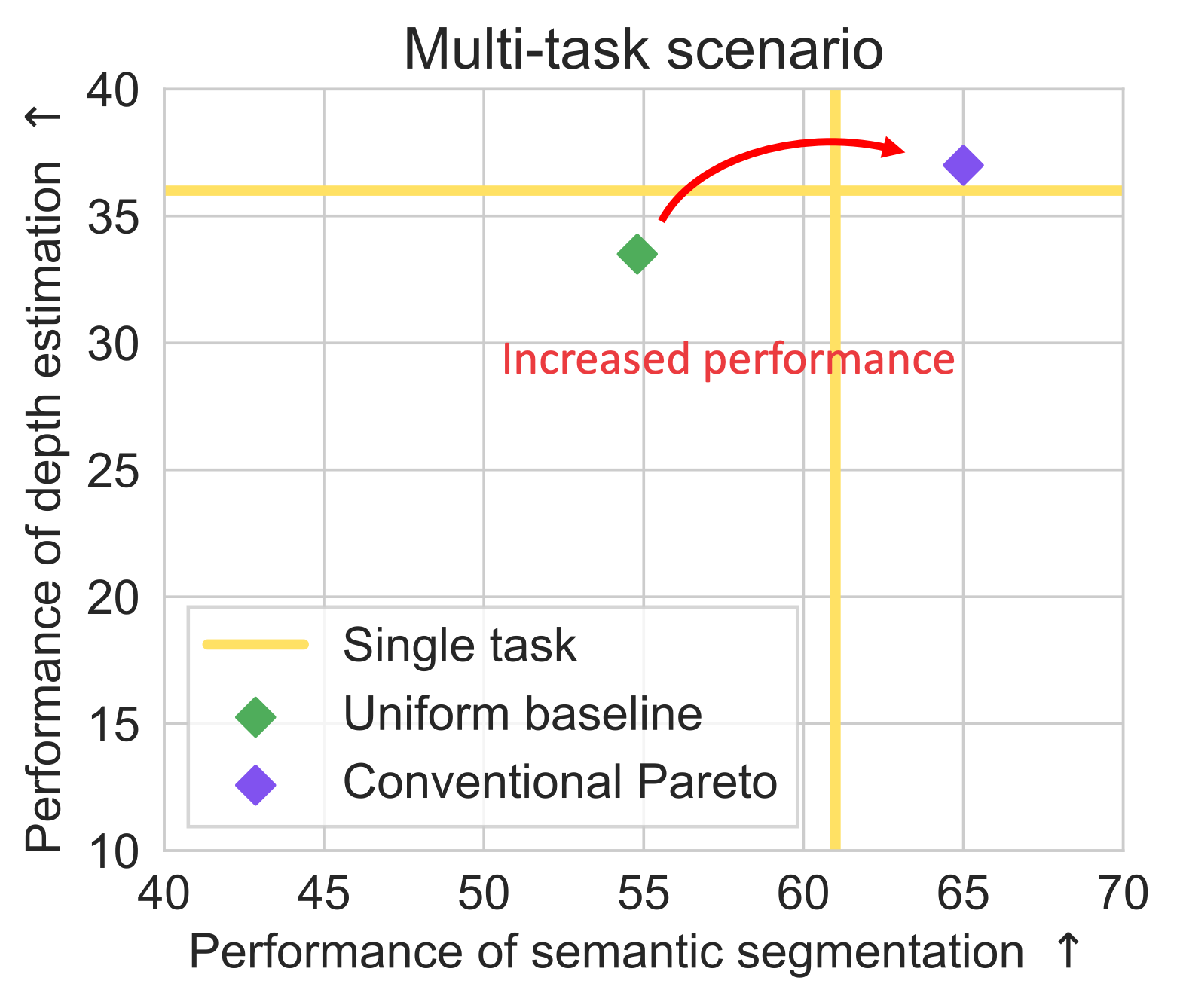}
			\caption{Cityscapes.}
			\label{fig:teaser-multitask}
	\end{subfigure}
	    \begin{subfigure}[t]{.23\textwidth}
			\centering
			\includegraphics[width=\textwidth]{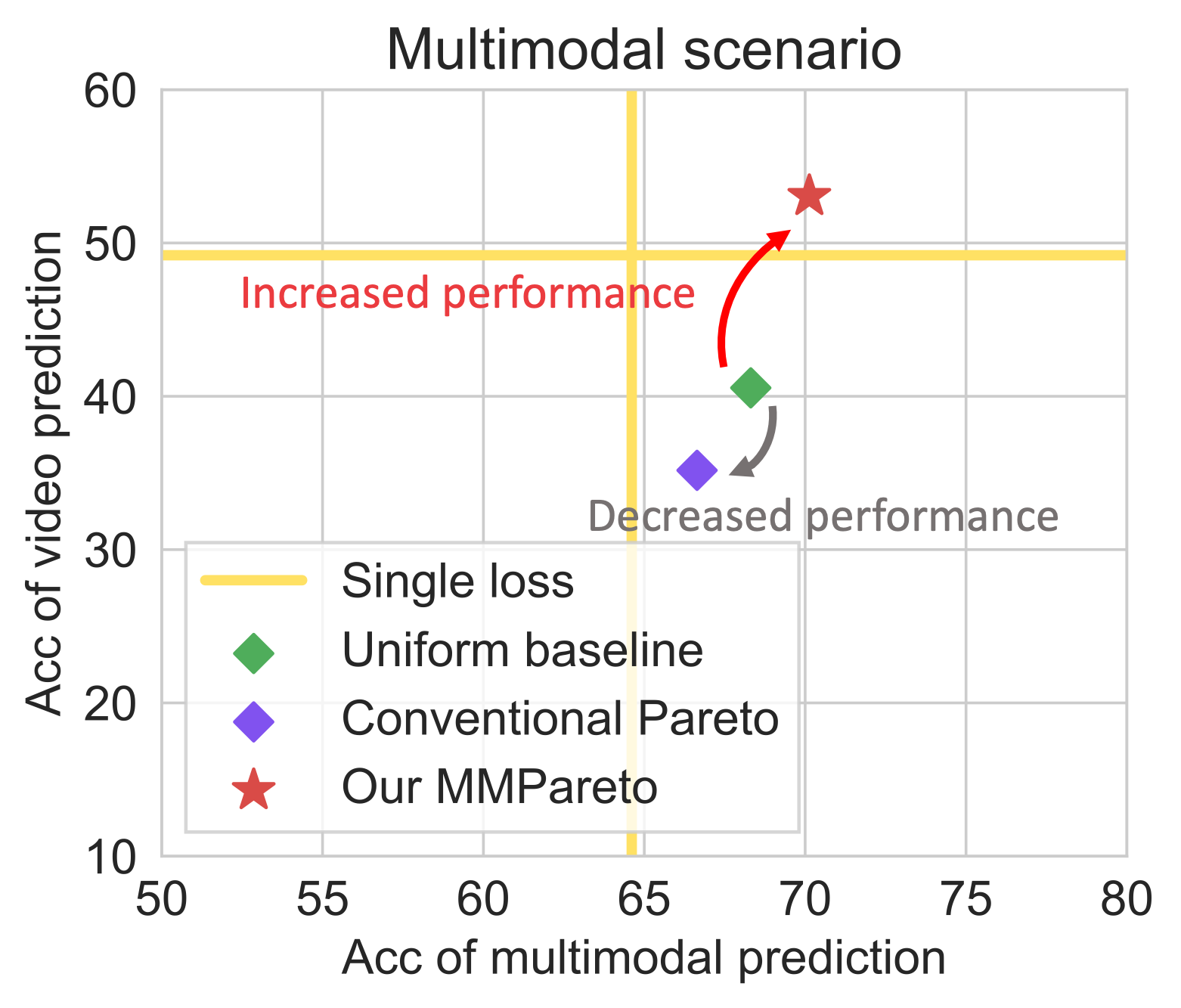}
			\caption{Kinetics Sounds.}
			\label{fig:teaser-multimodal}
	\end{subfigure}
	    \begin{subfigure}[t]{.23\textwidth}
			\centering
			\includegraphics[width=\textwidth]{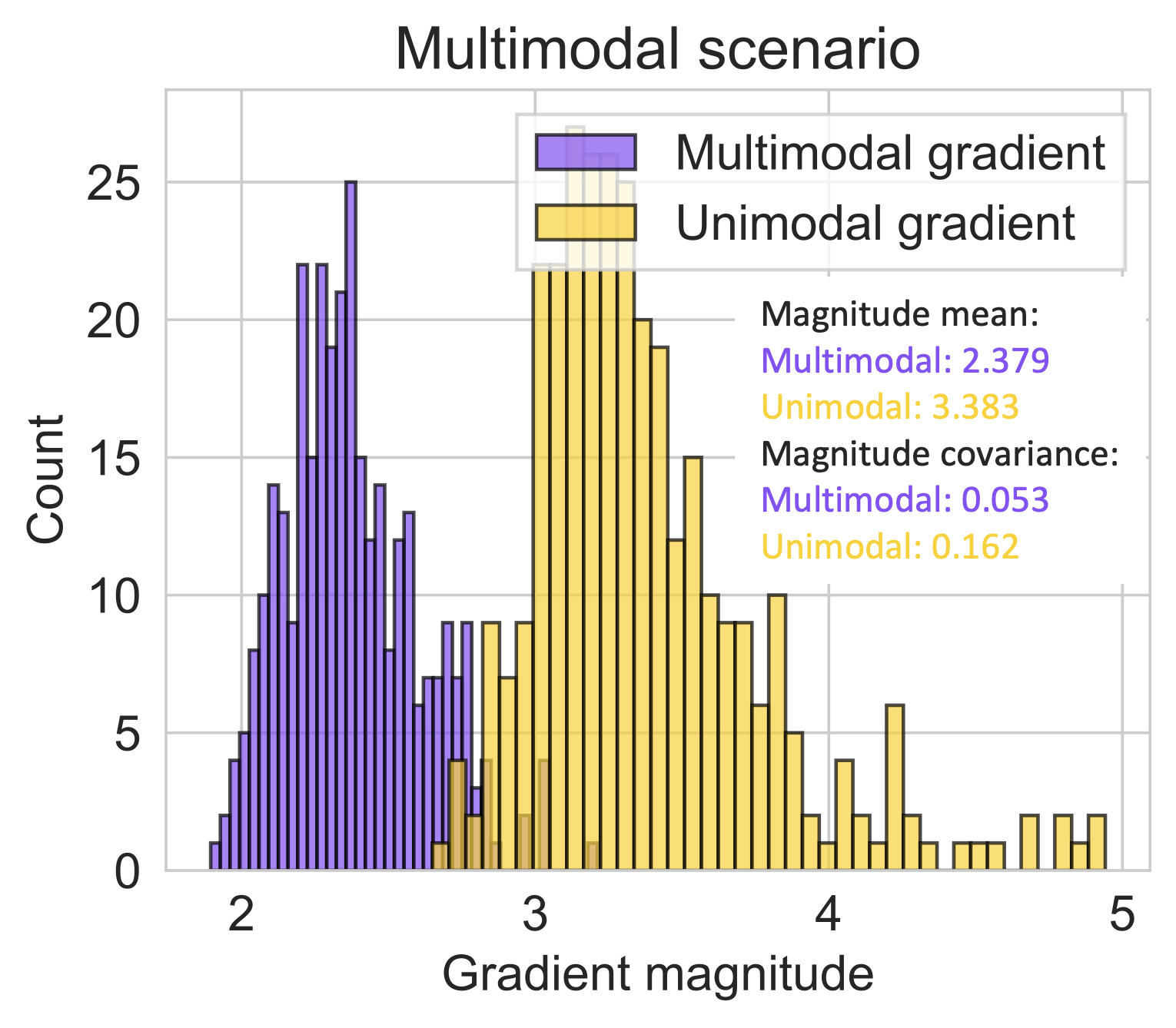}
			\caption{Kinetics Sounds.}
			\label{fig:teaser-multimodal-distribution}
	\end{subfigure}
 \vspace{-0.5em}
    \caption{\textbf{(a).} Cosine similarity between multimodal and unimodal gradients in the video encoder of Kinetics Sounds dataset~\citep{arandjelovic2017look}. \textbf{(b).} Methods performance on the multi-task dataset, Cityscapes~\citep{cordts2016cityscapes}. Results are from~\cite{sener2018multi} \textbf{(c).} Methods performance of multimodal and unimodal prediction in the video encoder of Kinetics Sounds. Single loss is the result of the individually trained model with one corresponding learning objective. \textbf{(d).} The gradient magnitude distribution for a fixed video encoder of Kinetics Sounds dataset. Each count is a mini-batch of SGD optimization. Uniform baseline is a basic way where all losses are equally summed without special integration.} 
    \vspace{-1em}
\end{figure*}

However, behind the effective performance, we observe a previously ignored risk in model optimization under this widely used multitask-like scenario, potentially limiting model ability. Every coin has two sides. Unimodal learning objectives undeniably effectively enhance the learning of corresponding modalities. Meanwhile, the optimization of parameters in unimodal encoder is influenced by both multimodal joint learning objective and its own unimodal learning objective. This entails the need to minimize two learning objectives concurrently, but usually, there does not exist a set of parameters that could satisfy this goal. Consequently, these multimodal and unimodal learning objectives could have conflict during optimization. In~\autoref{fig:teaser-conflic}, we take an example of the video encoder on the widely used Kinetics Sounds dataset. As the results, negative cosine similarity indicates that multimodal and unimodal gradients indeed have conflicts in direction during optimization. Especially, these conflicts at the early training stage could substantially harm the model ability~\citep{liu2020early}. Hence the primary multimodal learning is potentially disturbed.

To avoid optimization conflicts, it is essential to integrate both gradients well, making the unimodal gradient not affect the primary multimodal training but assist it. This necessity naturally accords with the idea of Pareto method~\citep{sener2018multi}, which aims to find a steep gradient direction that benefits all objectives and finally converges to a trade-off state of them. As~\autoref{fig:teaser-multitask}, on one of the representative multi-task cases,  semantic segmentation and depth estimation tasks on Cityscapes~\citep{cordts2016cityscapes} dataset, Pareto method has achieved ideal advancement in balancing the learning objective of these two tasks. Therefore, it is expected to keep superiority in solving conflicts in this multitask-like multimodal learning framework. However, the fact is contrary to the expectation. As~\autoref{fig:teaser-multimodal}, the conventional Pareto method~\citep{sener2018multi} surprisingly loses its efficacy, even is worse than the uniform baseline, where all gradients are equally summed.

To explore this counterintuitive phenomenon, we first analyze properties within multimodal learning scenarios. Unlike typical multi-task cases, multimodal joint loss is optimized with information from all modalities, while unimodal loss is optimized with information only from the corresponding modality. Accordingly, multimodal loss is naturally easier to learn, and verified with lower training error and faster convergence speed~\cite{wang2020makes}. We analyze the gradient magnitude distribution for a fixed set of parameters. As~\autoref{fig:teaser-multimodal-distribution}, the multimodal gradient magnitude is smaller than the unimodal one with a smaller batch sampling covariance. With these properties in gradient magnitude, we further theoretically analyze and empirically verify that the conventional Pareto could affect SGD noise strength and then bring the model to a sharper minima, weakening model generalization ability.

Based on the above analysis, it becomes imperative to well address gradient conflicts in multimodal scenarios. Hence, we propose the \textbf{M}ulti\textbf{M}odal \textbf{Pareto} (MMPareto) algorithm, which respectively takes the \emph{direction} and \emph{magnitude} into account during gradient integration. It ensures innocent unimodal assistance, where the final gradient is with direction common to all learning objectives and enhanced magnitude for improving generalization. We also provide an analysis of the method's convergence. Based on results across multiple types of datasets, our method effectively alleviates the imbalanced multimodal learning problem and could be well equipped with models with dense cross-modal interaction, like multimodal Transformers. As~\autoref{fig:teaser-multimodal}, our method provides both advanced multimodal performance and unimodal performance. What's more, the unimodal performance is even superior to the individually trained unimodal model, which was rarely achieved before. Moreover, we verify that the proposed method could also extend to multi-task cases with clear discrepancy in task difficulty, indicating its scalability.

Our contribution is three-fold. \textbf{Firstly,} we observe the previously ignored gradient conflict in the widely used multitask-like framework for the imbalanced multimodal learning problem. \textbf{Secondly,} we theoretically analyze the failure of Pareto method in multimodal case, and then propose the MMPareto algorithm which could provide innocent unimodal assistance with enhanced generalization. \textbf{Thirdly}, experiments across different datasets verify our theoretical analysis as well as superior performance.

\section{Related Work}
\subsection{Imbalanced Multimodal Learning}
Recent research has uncovered the imbalanced multimodal learning problem, as multimodal models tend to favor specific modalities, thereby constraining their overall performance~\citep{peng2022balanced,huang2022modality}. Several methods have been proposed for this problem, with a shared focus on targetedly enhancing optimization of each modality~\citep{wang2020makes,peng2022balanced,wu2022characterizing,fan2023pmr}. For example, \citet{xu2023mmcosine} focused on the fine-grained classification task, which has a higher demand for distinguishable feature distribution. \citet{yang2024Quantifying} put attention on the multimodal robustness. \citet{wei2024enhancing} introduced a Shapley-based sample-level modality valuation metric, to observe and alleviate the fine-grained modality discrepancy. Among them, multitask-like methods that directly incorporate targeted unimodal constraints have demonstrated superior effectiveness~\citep{wang2020makes,du2023uni,fan2023pmr}. However, under this multitask-like framework, optimization of unimodal encoder is simultaneously controlled by the multimodal joint learning objective and corresponding unimodal learning objective, which could cause gradient conflict, potentially harming the primary multimodal learning. In this paper, we observe and diminish the potential conflict by the proposed MMPareto algorithm. Our method could effectively alleviate the imbalanced multimodal learning problem, achieving considerable improvement.

\subsection{Pareto Integration in Multi-task Learning}
Shared parameter in multi-task learning is expected to fit several learning objectives simultaneously, resulting in the potential conflict problem during optimization. Hence, the Pareto method is introduced to integrate different gradients, finding a gradient common to all objectives and finally converging to a trade-off state of them~\citep{sener2018multi}. Besides, the idea of Pareto integration is extended from different perspectives, including more different trade-offs among different tasks~\citep{lin2019pareto} or faster convergence speed~\citep{ma2020efficient}, to better benefit multi-task learning. Similarly, we observe the optimization conflict in the shared unimodal encoder in the multitask-like multimodal framework. Inspired by the success of Pareto integration, we introduce this idea but surprisingly find it failed. We further analyze and find the harmed generalization of Pareto integration in multimodal scenarios, and then propose MMPareto algorithm, which could handle multimodal scenarios and multi-task cases with clear discrepancy in task difficulty.

\section{Method}

\subsection{Multitask-like Multimodal Framework}
\begin{figure}[t]
    \centering
    \includegraphics[width=1\linewidth]{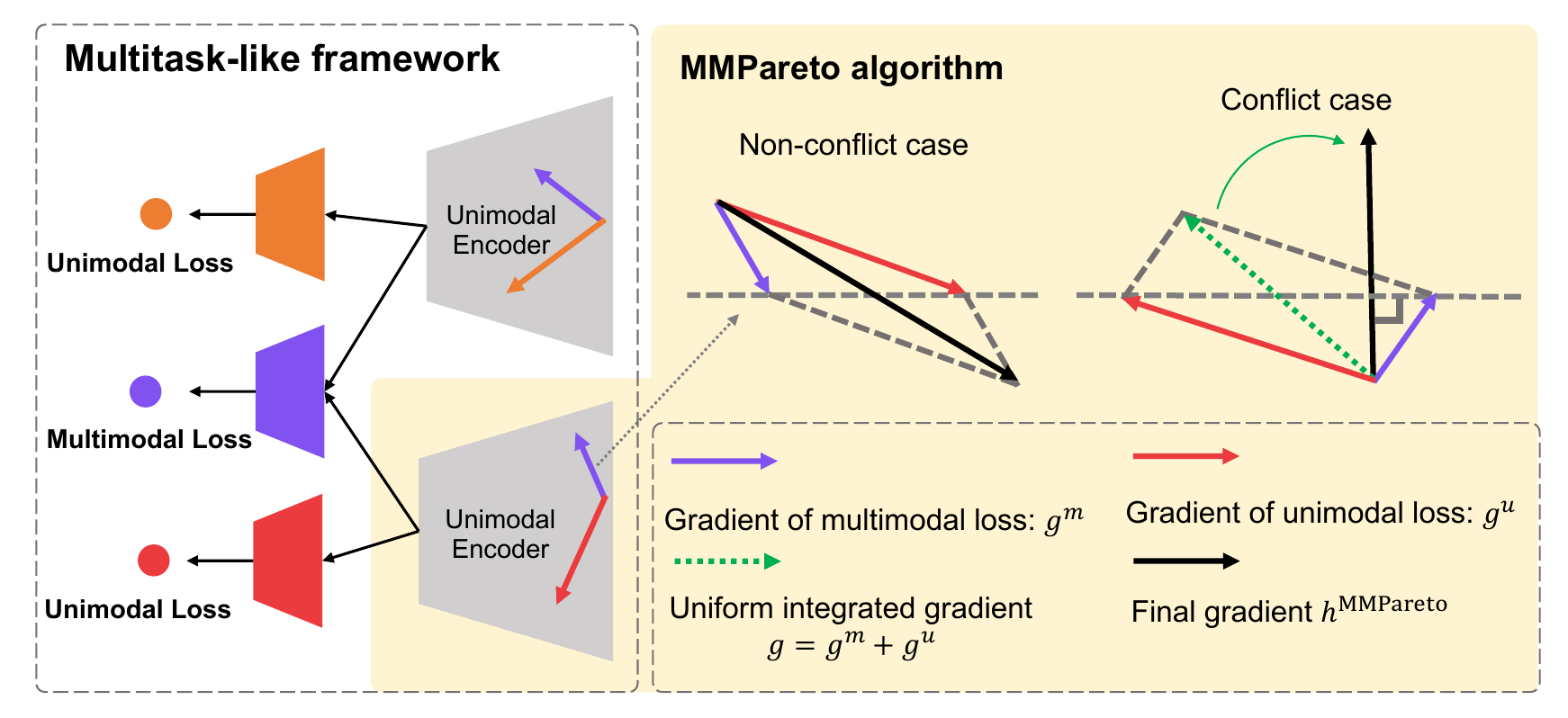}
    \vspace{-1.5em}
    \caption{Illustration of multimodal framework and gradient integration strategy of our MMPareto.}
    \vspace{-1em}
    \label{fig:method}
\end{figure}

In multimodal learning, models are expected to produce correct predictions by integrating information from multiple modalities. Therefore, there are often \emph{multimodal joint loss}, which takes the prediction of fused multimodal feature. However, only utilizing such joint loss to optimize all modalities together could result in the optimization process being dominated by one modality, leaving others being severely under-optimized~\citep{peng2022balanced,huang2022modality}. To overcome this imbalanced multimodal learning problem, introducing \emph{unimodal loss} which targets the optimization of each modality is widely used and verified effective for alleviating this imbalanced multimodal learning problem~\citep{wang2020makes}. In these scenarios, the loss functions are:
\begin{equation}
\label{equ:loss}
    \gL= \gL_{m}+ \sum^n_{k=1} \gL_{u}^k,
\end{equation}
where $\gL_{m}$ is the multimodal joint loss and $\gL_{u}^k$ is the unimodal loss for modality $k$. $n$ is the number of modalities. We mainly consider the multimodal discriminative task, and all losses are cross-entropy loss functions. The illustration of this multitask-like multimodal framework is shown in the left part of~\autoref{fig:method}.

\subsection{SGD Property and Hypothesis}
\label{sec:hypothesis}
In SGD optimization, based on former studies~\citep{jastrzkebski2017three}, for an arbitrary loss $\gL$, when the batch size is sufficiently large, with the central limit theorem, the gradient of parameters $\theta$ at $t-$th mini-batch $S$, $\rvg_{S}(\theta(t)) =\frac{1}{|S|} \sum_{i=1}^{|S|} \nabla_{\theta(t)} \gL\left(X_i, Y_i\right)$, is unbiased estimations of full gradient, $\rvg_{N}(\theta(t)) =\frac{1}{N} \sum_{i=1}^{N} \nabla_{\theta(t)} \gL\left(X_i, Y_i\right)$:
\begin{equation}
\begin{gathered}
{\rvg}_S(\theta(t)) \sim \mathcal{N}\left(\rvg_N(\theta(t)), \frac{1}{|S|} C\right).\\
\end{gathered}
\end{equation}

In other words, ${\rvg}_S(\theta(t))$ is a random variable with covariance $\frac{1}{|S|} C$. $C$ is brought by random batch sampling. $N$ is the number of training samples. $(X_i, Y_i)$ is a sample. The full gradient is calculated based on all training samples. 

As~\autoref{equ:loss}, multimodal framework has both multimodal loss function and unimodal loss function. For $\theta^{k}$, the unimodal encoder parameter of modality $k$, gradients of $\gL_{m}$ and $\gL_{u}^k$ at iteration $t$ satisfy:
\begin{gather}
\label{equ:guasian_m}
{\rvg}^m_S(\theta^k(t)) \sim \mathcal{N}\left(\rvg_N^m(\theta^k(t)), \frac{1}{|S|} C^m\right), \\
\label{equ:guasian_u}
{\rvg}^u_S(\theta^k(t)) \sim \mathcal{N}\left(\rvg_N^u(\theta^k(t)), \frac{1}{|S|} C^u\right),
\end{gather}
where $\frac{1}{|S|} C^m$ and $\frac{1}{|S|} C^u$ are the batch sampling covariance. 

In multimodal cases, unimodal loss only receives the prediction based on data of the corresponding modality. In contrast, the multimodal loss is optimized with more sufficient information from data of all modalities, making it easier to be trained. It has been verified that multimodal loss tends to converge with faster speed and with lower training error than unimodal one~\cite{wang2020makes}. 

Beyond former studies, we also analyze the gradient magnitude distribution for a fixed set of parameters across different datasets. Results are as~\autoref{fig:teaser-multimodal-distribution} and~\autoref{fig:cd-norm-visual}. Each count is a different mini-batch. Firstly, multimodal gradients have a smaller magnitude than unimodal gradients, which could be brought by the lower multimodal error~\cite{chen2018gradnorm}. Besides the magnitude, the batch sampling covariance $\frac{1}{|S|} C$ of gradient variables also differs between multimodal and unimodal ones. Multimodal gradients have a smaller covariance. The reason could be that more discriminative information from all modalities is more deterministic for classification, which aligns with the motivation of multimodal learning that introduces more modalities to reduce uncertainty~\cite{trick2019multimodal}. Then the multimodal training error has a less variants across different batches. Overall, based on former studies and our verification, we could conclude properties of multimodal and unimodal gradients, and have~\autoref{hyposis}.

\begin{hypothesis} 
\label{hyposis}
In multitask-like multimodal cases, for the shared unimodal encoder, the gradient of unimodal loss tends to have a larger magnitude and larger batch sampling covariance than easy-to-learnt multimodal loss.
\end{hypothesis}

\subsection{Pareto Integration in Multimodal Learning}

\subsubsection{Pareto Integration}

In multimodal cases, multimodal loss and unimodal loss are tightly related, but their gradients may still exist conflicts, as~\autoref{fig:teaser-conflic}. Hence, how to well integrate ${\rvg}^m_S(\theta^k(t))$ and ${\rvg}^u_S(\theta^k(t))$ needs to be solved. This accords with the idea of Pareto method in multi-task learning~\citep{sener2018multi}. In Pareto method, at each iteration, gradients are assigned different weights, and the weighted combination is the final gradient, which can provide descent direction that benefits all learning objectives. Finally, parameters can converge to a trade-off state, Pareto-optimality, in which no objective can be advanced without harming any other objectives. It is natural to introduce Pareto integration into multimodal framework, avoiding conflict between multimodal and unimodal gradients. Concretely, in our case, for modality $k$, the Pareto algorithm is formulated to solve:
\begin{equation}
\label{equ:pareto}
\begin{gathered}
\min_{\alpha^m,\alpha^u \in \gR} {\| \alpha^m \rvg^m_S +\alpha^u \rvg^u_S \|}^2 \\
s.t. \quad \alpha^m,\alpha^u \geq 0, \alpha^m+\alpha^u=1,
\end{gathered}
\end{equation}
where $\| \cdot \|$ denotes the $L_2$-norm. For brevity, we denote $\{\rvg^i_S(\theta^k(t))\}_{i \in \{m,u\}}$ as $\{\rvg^i_S\}_{i \in \{m,u\}}$ for modality $k$ in some part. This problem is equal to finding the minimum-norm in the convex hull of the family of gradient vectors $\{\rvg^i_S\}_{i \in \{m,u\}}$. \cite{desideri2012multiple} showed that either found minimum-norm to this optimization problem is $0$ and the corresponding parameters are Pareto-stationary which is a necessary condition for Pareto-optimality, or it can provide descent direction common to all learning objectives.

\subsubsection{Analytic Solution of Pareto}
As results in~\autoref{fig:teaser-multimodal}, Pareto integration is expected to exhibit its advantage under multitask-like multimodal framework but surprisingly fails. To explore the hidden reason, we further analyze the property of the Pareto integration method. Based on~\autoref{hyposis}, multimodal gradient tends to with a smaller magnitude than that of unimodal one, \emph{i.e.,} $\| \rvg^m_S\| <  \| \rvg^u_S \|$. Then, for the optimization problem of~\autoref{equ:pareto}, we can have its analytic solution:
\begin{gather*}
\begin{cases}
\alpha^m =1, \alpha^u=0 \quad & \cos\beta \geq \frac{\| \rvg^m_S \|}{\| \rvg^u_S \|}, \\
\alpha^m=\frac{(\rvg^u_S-\rvg^m_S)^\top\rvg^u_S}{{\| \rvg^m_S-\rvg^u_S \|}^2}, \alpha^u=1-\alpha^m & \text{otherwise},
\end{cases} 
\end{gather*}
where $\beta$ is the angle between $\rvg^m_S$ and $\rvg^u_S$. 
Here we further analyze the above Pareto analytic solution. When $\cos\beta \geq \frac{\| \rvg^m_S \|}{\| \rvg^u_S \|}$, we have $\alpha^m > \alpha^u$. Otherwise, we also have:
\begin{align*}
&\alpha^m - \alpha^u \\
& = \frac{(\rvg^u_S-\rvg^m_S)^\top\rvg^u_S}{{\| \rvg^m_S-\rvg^u_S \|}^2}  - (1- \frac{(\rvg^u_S-\rvg^m_S)^\top\rvg^u_S}{{\| \rvg^m_S-\rvg^u_S \|}^2})  \\
& = \frac{ {\|\rvg^u_S \|}^2 - \| \rvg^u_S\| \|\rvg^m_S \|   \cos\beta   }{{\| \rvg^m_S-\rvg^u_S \|}^2} - \frac{ {\|\rvg^m_S \|}^2 - \| \rvg^u_S\| \|\rvg^m_S \|   \cos\beta   }{{\| \rvg^m_S-\rvg^u_S \|}^2} \\
& > 0. \quad \text{($\| \rvg^m_S \| <  \| \rvg^u_S \|$)}
\end{align*}

\begin{remark} 
\label{remark:short-prior}
The conventional Pareto method would assign a larger weight to the multimodal gradient with a smaller magnitude.
\end{remark}

Overall, as stated in~\autoref{remark:short-prior}, we can conclude that the Pareto method tends to assign larger weight to the multimodal gradient $\rvg^m_S$ during integration, \emph{i.e.,} $\alpha^m>\frac{1}{2}$.

\subsubsection{Generalization Harmed Risk of Pareto}
In this section, we follow the notations of~\autoref{sec:hypothesis}. During training, gradients of multiple losses are calculated separately, so they can be treated as being independent~\cite{fan2022maxgnr}. Then, when without any gradient integration strategy \emph{i.e.,} uniform baseline where all losses are equally summed, the final gradient is ${\rvh}_S(\theta^k(t)) = {\rvg}^m_S(\theta^k(t)) + {\rvg}^u_S(\theta^k(t))$. And based on~\autoref{equ:guasian_m} and~\autoref{equ:guasian_u}, it satisfies:
\begin{equation}
{\rvh}_S(\theta^k(t))  \sim \mathcal{N}\left( \rvg^m_N(\theta^k(t))+\rvg^u_N(\theta^k(t)), \frac{  C^m +   C^u}{|S|} \right).
\end{equation}

Then, use the final gradient ${\rvh}_S(\theta^k(t))$ to update $\theta^k$:
\begin{align}
\theta^k(t+1)& =\theta^k(t)-\eta {\rvh}_S(\theta^k(t)), \\
\label{equ:uniform-noise}
& =\theta^k(t)-\eta{\rvh}_N(\theta^k(t)) +\eta \epsilon_t,
\end{align}
where ${\rvh}_N(\theta^k(t))=\rvg^m_N(\theta^k(t))+ \rvg^u_N(\theta^k(t))$ and $\eta>0$ is the learning rate. $\epsilon_t \sim \mathcal{N}\left(0, \frac{ C^m +  C^u}{|S|} \right)$ is often considered as the noise term of SGD~\citep{zhu2018anisotropic}.

Then, we further consider the SGD optimization with conventional Pareto gradient integration. We know that at each iteration, Pareto method would return weight $\alpha^m$ and $\alpha^u$ for the integration of ${\rvg}^m_S(\theta^k(t))$ and ${\rvg}^u_S(\theta^k(t))$. Then, we have the final gradient after integration: ${\rvh^{\text{Pareto}}_S}(\theta^k(t))=2\alpha^m  {\rvg}^m_S(\theta^k(t)) + 2\alpha^u {\rvg}^u_S(\theta^k(t))$, which satisfies:
\begin{equation}
\label{equ:h_pareto}
{\rvh^{\text{Pareto}}_S}(\theta^k(t))  \sim \mathcal{N}\left( {\rvh^{\text{Pareto}}_N}(\theta^k(t)), \frac{ (2\alpha^m)^2 C^m +  (2\alpha^u)^2 C^u}{|S|} \right), 
\end{equation}
where ${\rvh^{\text{Pareto}}_N}(\theta^k(t))=2\alpha^m  {\rvg}^m_N(\theta^k(t)) + 2\alpha^u {\rvg}^u_N(\theta^k(t))$.
Here we use $2\alpha^i$ as the gradient weight to keep the same gradient weight summation with uniform baseline (\emph{i.e., $2\alpha^m+2\alpha^u =2$}). Then, when using Pareto integration, the parameter is updated as:
\begin{align}
\theta^k(t+1) & =\theta^k(t)-\eta {\rvh^{\text{Pareto}}_S}(\theta^k(t)) \\
\label{equ:pareto-noise}
& =\theta^k(t)-\eta{\rvh^{\text{Pareto}}_N}(\theta^k(t)) +\eta\zeta_t.
\end{align}
$\zeta_t \sim \mathcal{N}\left(0,  \frac{ (2\alpha^m)^2 C^m +  (2\alpha^u)^2 C^u}{|S|} \right)$ is the SGD noise term. 

Based on~\autoref{hyposis}, the unimodal gradient tends to have a larger batch sampling covariance. Suppose the covariance of multimodal gradient and unimodal gradient satisfies $ kC^m =C^u  $, where $k>1$. Then, we can explore relation between $\zeta_t$ and $\epsilon_t$. When the covariance of $\zeta_t $ is less than that of $ \epsilon_t$, it should satisfy:
\begin{gather}
(2\alpha^m)^2 C^m +  (2\alpha^u)^2 C^u  < C^m +C^u \\
 (2\alpha^m)^2 C^m + (2(1-\alpha^m))^2 \cdot k C^m < (k+1) C^m \\
 (2\alpha^m)^2+4k-8\alpha^mk+(2\alpha^m)^2k < k+1 \\
 (\alpha^m - \frac{1}{2})((4k+4)\alpha^m+2-6k) <0
\end{gather}
Hence when $ \frac{1}{2}< \alpha^m < \frac{3k-1}{2k+2}$, the covariance of $\zeta_t $ is less than that of $ \epsilon_t$, \emph{i.e.,} $\text{var}(\zeta_t)<\text{var}(\epsilon_t)$. In addition, based on~\autoref{remark:short-prior}, Pareto weight satisfies $\frac{1}{2}<\alpha^m \leq 1$. Combine with the value range of $\alpha^m$, then:

\begin{itemize}
\vspace{-0.5em}  \item For case $k\leq 3$, when $ \frac{1}{2}< \alpha^m < \frac{3k-1}{2k+2}$, it has $\text{var}(\zeta_t)<\text{var}(\epsilon_t)$;
\vspace{-0.5em}
  \noindent  \item For case $k>3$, it has $\frac{3k-1}{2k+2} >1$,  $\text{var}(\zeta_t)<\text{var}(\epsilon_t)$ holds for all Pareto weight $\alpha^m$.
\end{itemize}

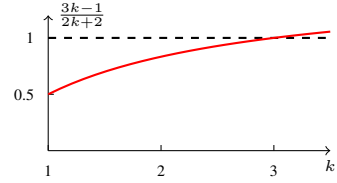
\begin{figure}
\centering
\begin{tikzpicture}[scale=1.5]
\draw[->](1,0)--(3.5,0)node[left,below,font=\tiny]{$k$};
\draw[->](1,0)--(1,1.2)node[right,font=\tiny]{$\frac{3k-1}{2k+2}$};

\draw[thick][dashed](1,1)--(3.5,1);

\foreach \x in {1,2,3}{\draw(\x,0)--(\x,0.05)node[below,outer sep=2pt,font=\tiny]at(\x,0){\x};}

\foreach \y in {0.5,1}{\draw(1,\y)--(1.05,\y)node[left,outer sep=2pt,font=\tiny]at(1,\y){\y};}

\draw[thick][color=red  ,domain=1:3.5]plot(\x,{(3*\x-1)/(2*\x+2)});
\end{tikzpicture}
\vspace{-1em}
\caption{Value of $\frac{3k-1}{2k+2}$ varies with $k$.}
\label{fig:k}
\vspace{-1em}
\end{figure}

Also, in~\autoref{fig:k}, we give the value of $\frac{3k-1}{2k+2}$ varies with $k$. When $k\leq3$, it can observe a rapid increase of $\frac{3k-1}{2k+2}$ when the value of $k$ increases. In addition, as shown in~\autoref{fig:teaser-multimodal-distribution} and~\autoref{fig:cd-norm-visual}, the difference of covariance between multimodal and unimodal gradient are often more than 3 times in experiments. These phenomena indicate that $\text{var}(\zeta_t)<\text{var}(\epsilon_t)$ could often happen in practice.

According to existing studies, the strength of noise term influences the minima found by SGD, and larger noise term strength tends to bring flatter minima, which typically generalize well~\citep{hochreiter1997flat,keskar2016large,neyshabur2017exploring}. Therefore, when $\text{var}(\zeta_t)<\text{var}(\epsilon_t)$, SGD noise term of conventional Pareto integration in~\autoref{equ:pareto-noise} can with smaller strength, compared with SGD noise term of uniform baseline in~\autoref{equ:uniform-noise}. Then it would bring model with sharper minima, and further harm the model generalization ability.

\subsection{Multimodal Pareto Algorithm}
Based on the above analysis, conventional Pareto method could result in a sharper minima and then weakening model generalization in multimodal learning. It becomes essential to well address gradient conflicts in multimodal scenarios. In this paper, we propose the \textbf{M}ulti\textbf{M}odal \textbf{Pareto} (MMPareto) algorithm. It considers both the \emph{direction} and \emph{magnitude} during gradient integration, to provide innocent unimodal assistance where the final gradient is with a direction common to all learning objectives while an enhanced magnitude for improving generalization. Our method considers the conflict case and non-conflict case respectively. The overall algorithm is shown in Algorithm~\ref{alg:mmpareto} and illustrated in~\autoref{fig:method}. In the following part, we take the encoder of modality $k$ for an example, and encoders of all modalities follow the same integration. We also omit $\theta^k(t)$ for brevity.

\textbf{Non-conflict case.} 
We first consider the case $\cos\beta \geq 0$. Under this case, the cosine similarity between $\rvg^m_S$ and $\rvg^u_S$ is positive. For the direction, the arbitrary convex combination of the family of gradient vectors $\{\rvg^i_S\}_{i \in \{m,u\}}$ is common to all learning objectives. Therefore, we assign $2\alpha^m = 2\alpha^u = 1$ instead of analytic solution of Pareto during integration in this case, to have enhanced SGD noise term. With this setting, our final gradient $\rvh^{\text{MMPareto}}_S  \sim \mathcal{N}\left( \rvg^m_N+\rvg^u_N, \frac{  C^m +   C^u}{|S|} \right)$. The noise term is $\xi_t \sim \mathcal{N}\left(0, \frac{ C^m +  C^u}{|S|} \right)$ with the enhanced strength, compared with conventional Pareto noise term $\zeta_t$ in~\autoref{equ:pareto-noise}.

\textbf{Conflict case.} For the case $\cos\beta < 0$, it is essential to find the direction that is common to all losses and enhance the SGD noise strength during gradient integration. Hence we first solve the Pareto optimization problem of~\autoref{equ:pareto}, obtaining $\alpha^m$ and $\alpha^u$, which could provide a non-conflict direction. Furthermore, to enhance the strength of noise term, we increase the magnitude of the final gradient. The magnitude of uniform baseline is used as the benchmark, to adjust in a proper range:
\begin{equation}
\rvh^{\text{MMPareto}}_S = \frac{2\alpha^m \rvg^m_S +2\alpha^u \rvg^u_S}{\| 2\alpha^m \rvg^m_S +2\alpha^u \rvg^u_S\|} \cdot \| \rvg^m_S+\rvg^u_S \|.
\end{equation}
In this case, the final gradient is $\rvh^{\text{MMPareto}}_S = \lambda \rvh^{\text{Pareto}}_S $, where $\lambda = \| \rvg^m_S+\rvg^u_S \| / \| 2\alpha^m \rvg^m_S +2\alpha^u \rvg^u_S \|$. Based on~\autoref{remark:short-prior}, the smaller multimodal magnitude is with larger weight, we can have that $\|2\alpha^m \rvg^m_S +2\alpha^u \rvg^u_S \|<  \| \rvg^m_S+\rvg^u_S \|$, and accordingly $\lambda >1$. Then, the final gradient satisfies: $\rvh^{\text{MMPareto}}_S  \sim \mathcal{N}\left( \lambda \rvh^{\text{Pareto}}_N, \lambda^2 \cdot \frac{ (2\alpha^m)^2 C^m +  (2\alpha^u)^2 C^u}{|S|} \right)$. The noise term is $\xi_t \sim \mathcal{N}\left(0, \lambda^2 \cdot \frac{ (2\alpha^m)^2 C^m +  (2\alpha^u)^2 C^u}{|S|} \right)$, which has a larger strength than noise term $\zeta_t$ of conventional Pareto in~\autoref{equ:pareto-noise}. The noise strength is enhanced, and model generalization is accordingly improved.

After integration, to further enhance the generalization, we slightly increase the magnitude of the final gradient: $\rvh^{\text{MMPareto}}_S = \gamma \rvh^{\text{MMPareto}}_S$, where $\gamma>1$ is the rescale factor.

\begin{algorithm}[t]
\caption{MMPareto}
\label{alg:mmpareto}
\begin{algorithmic}
\STATE {\bfseries Require:}Training dataset $\mathcal{D}$, iteration number $T$, initialized unimodal encoder parameters $\theta^{k}$, $k \in \{1,2,\cdots,n\}$, other parameters $\theta^{\text{other}}$.
\FOR{$t=0,\cdots,T-1$} 
    \STATE Sample a fresh mini-batch $S$ from $\mathcal{D}$;
    \STATE Feed-forward the batched data $S$ to the model;
    \STATE Calculate gradient using back-propagation;
    \STATE Update $\theta^{\text{other}}$ without gradient integration method;
    \FOR{$k=1,\cdots,n$}
        \STATE Obtain $\rvg^m_S$ and $\rvg^u_S$ for $k$-th unimodal encoder;
        \STATE Calculate $\cos\beta$; $\beta$ is angle between $\rvg^m_S$ and $\rvg^u_S$;
        \STATE Solve problem of~\autoref{equ:pareto}, obtain $\alpha^m$, $\alpha^u$;
        \IF{$\| \alpha^m \rvg^m_S +\alpha^u \rvg^u_S \| =0$}
            \STATE Find the Pareto stationarity;
        \ENDIF
        \IF{$\cos\beta \geq 0$}
            \STATE  $2\alpha^m=2\alpha^u=1$;
        \ENDIF
        \STATE Integrate gradient: $\rvh_S^{\prime}=2\alpha^m \rvg^m_S +2\alpha^u \rvg^u_S$;
        \STATE $\rvh_S^{\text{MMPareto}} = \underbrace{{\rvh_S^{\prime}}/{\| \rvh_S^{\prime} \|}}_{\text{Keep non-conflict direction}} \cdot  \underbrace{\gamma \|\rvg^m_S +\rvg^u_S\|}_{\text{Enhanced magnitude}}$;
        \STATE Update $\theta^k$ with ${\rvh}_S^{\text{MMPareto}}$.
    \ENDFOR
\ENDFOR
\end{algorithmic}
\end{algorithm}

Overall, MMPareto provides both non-conflict direction and enhanced SGD noise strength, helping the model converge to a flatter minima and generalize better. Beyond that, we also analyze the convergence of proposed MMPareto method. As~\autoref{the:converge}, our algorithm is guaranteed to converge to a Pareto stationarity. Detailed proof and related experiments are provided in~\autoref{sec:convergence} and~\autoref{sec:converge-exp}.

\begin{remark}
\label{the:converge}
The proposed MMPareto method admits an iteration sequence that converges to a Pareto stationarity.
\end{remark}

\section{Experiment}

\subsection{Dataset and Experiment Settings}

\textbf{CREMA-D}~\citep{cao2014crema} is an audio-visual dataset for emotion recognition, covering 6 usual emotions. \textbf{Kinetics Sounds}~\citep{arandjelovic2017look} is an audio-visual dataset containing 31 human action classes. \textbf{Colored-and-gray-MNIST}~\citep{kim2019learning} is a synthetic dataset based on MNIST~\citep{lecun1998gradient}. Each instance contains two kinds of images, a gray-scale and a monochromatic colored image. \textbf{ModelNet40}~\citep{wu20153d} is a dataset with 3D objects, covering 40 categories. This dataset could be used to classify 3D objects based on the multiple 2D views data~\citep{su2015multi}.

\begin{figure}[t]
\centering
	\begin{subfigure}[t]{.22\textwidth}
			\centering
			\includegraphics[width=\textwidth]{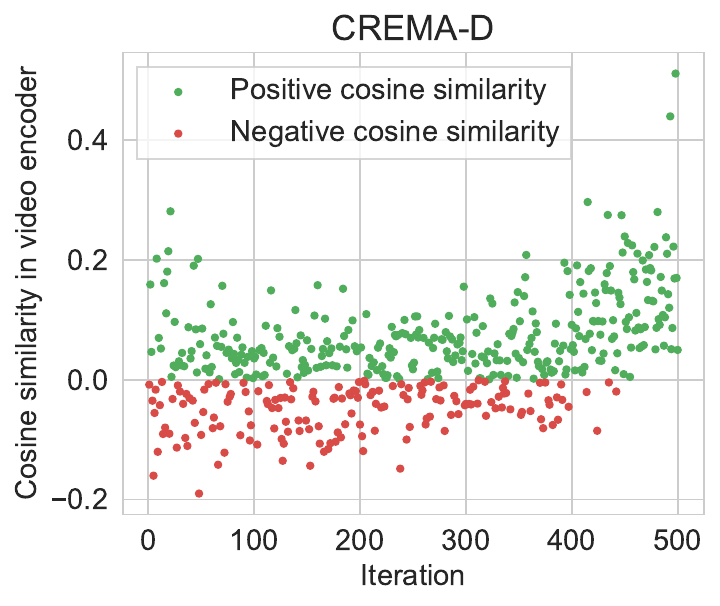}
			\caption{Direction conflict.}
			\label{fig:cd-conflic-visual}
	\end{subfigure}
    \begin{subfigure}[t]{.22\textwidth}
			\centering
			\includegraphics[width=\textwidth]{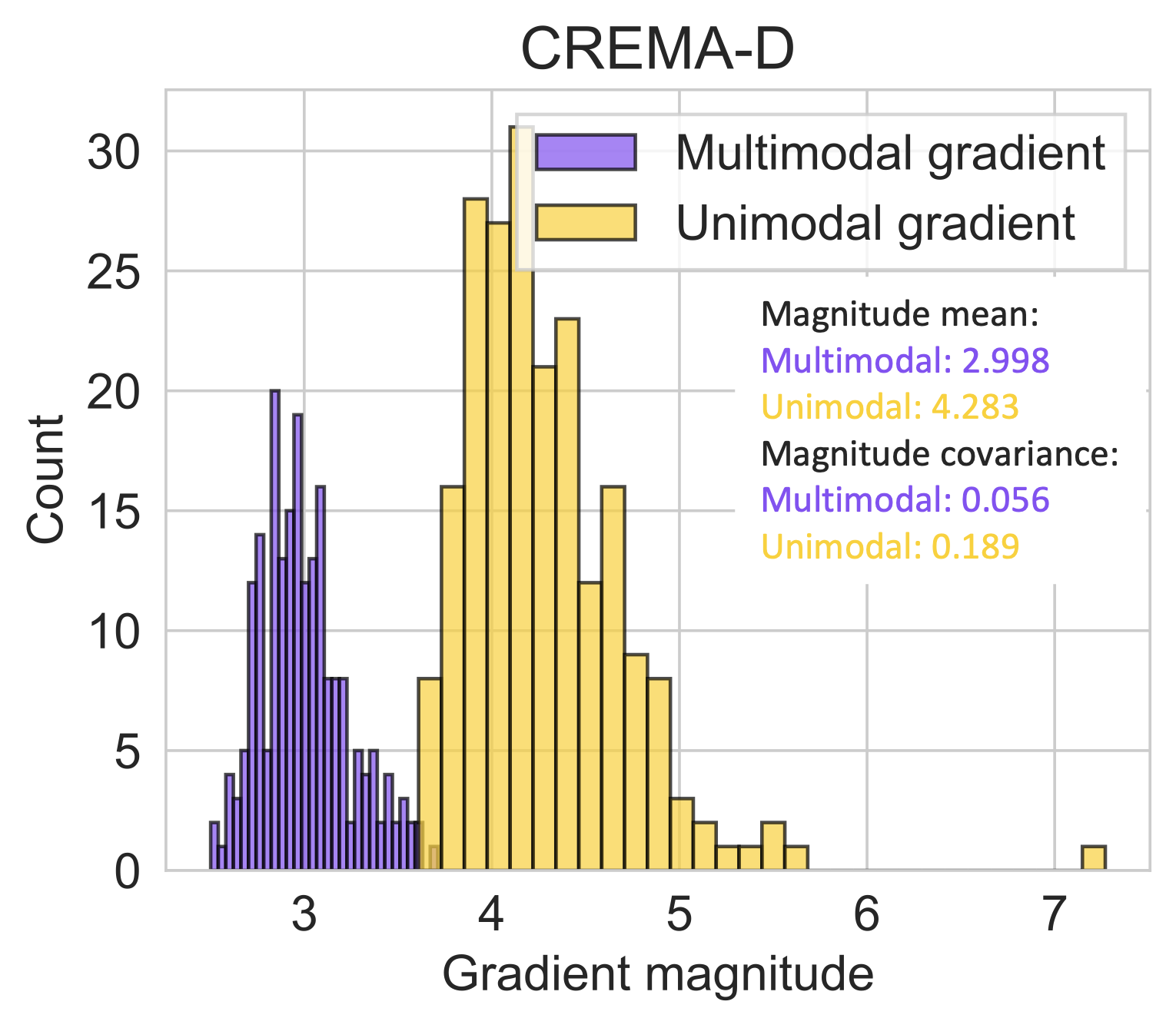}
			\caption{Gradient distribution.}
			\label{fig:cd-norm-visual}
	\end{subfigure}
 \vspace{-0.5em}
    \caption{\textbf{(a).} Cosine similarity between gradients of multimodal and unimodal loss in the video encoder of CREMA-D. \textbf{(b).} Gradient magnitude distribution in the video encoder of CREMA-D.}
     \vspace{-1.5em}
    \label{fig:cd}
\end{figure}

When not specified, ResNet-18~\citep{he2016deep} is used as the backbone in experiments and models are trained from scratch. Unimodal modal features are integrated with late fusion method. Specifically, for the Colored-and-gray MNIST dataset, we build a neural network with 4 convolution layers and 1 average pool layer
as the encoder, like~\cite{fan2023pmr} does. During the training, we use SGD with momentum ($0.9$) and $\gamma=1.5$ in experiments. More details are provided in~\autoref{sec:dataset}.

\begin{figure*}[t]
\centering
            \begin{subfigure}[t]{.23\textwidth}
			\centering
			\includegraphics[width=\textwidth]{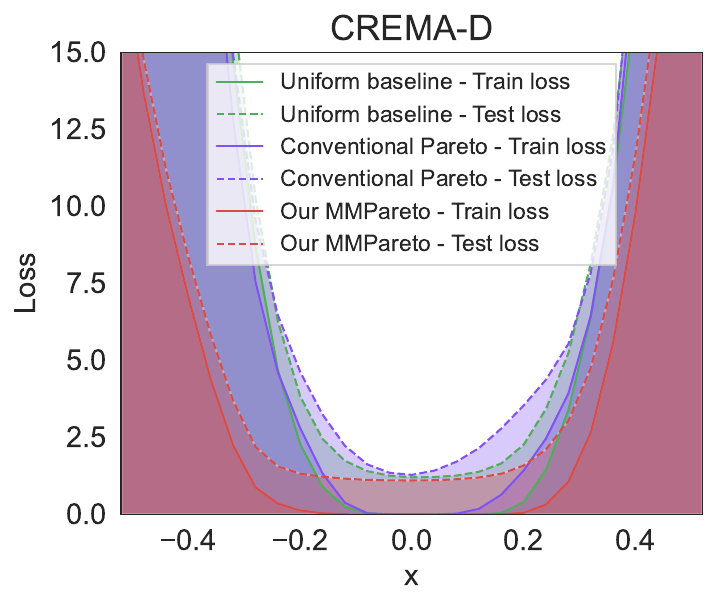}
			\caption{}
			\label{fig:cd-1d-loss}
	\end{subfigure}
	    \begin{subfigure}[t]{.23\textwidth}
			\centering
			\includegraphics[width=\textwidth]{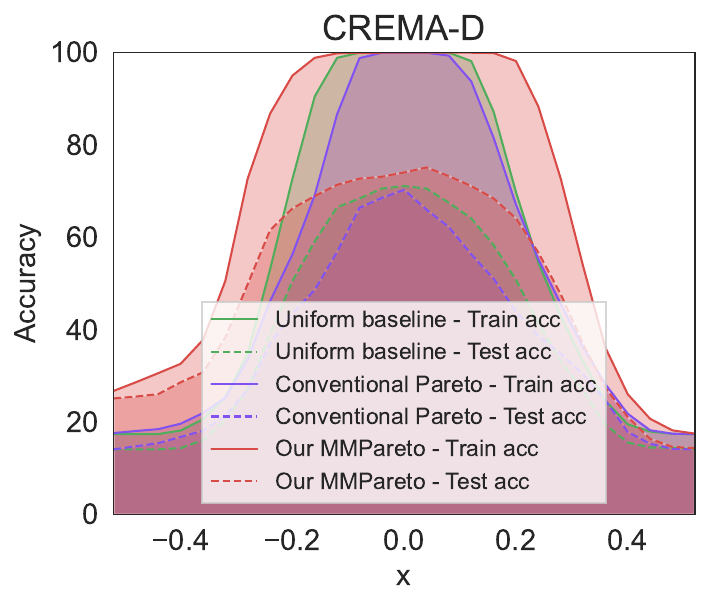}
			\caption{}
			\label{fig:cd-1d-acc}
	\end{subfigure}
            \begin{subfigure}[t]{.23\textwidth}
			\centering
			\includegraphics[width=\textwidth]{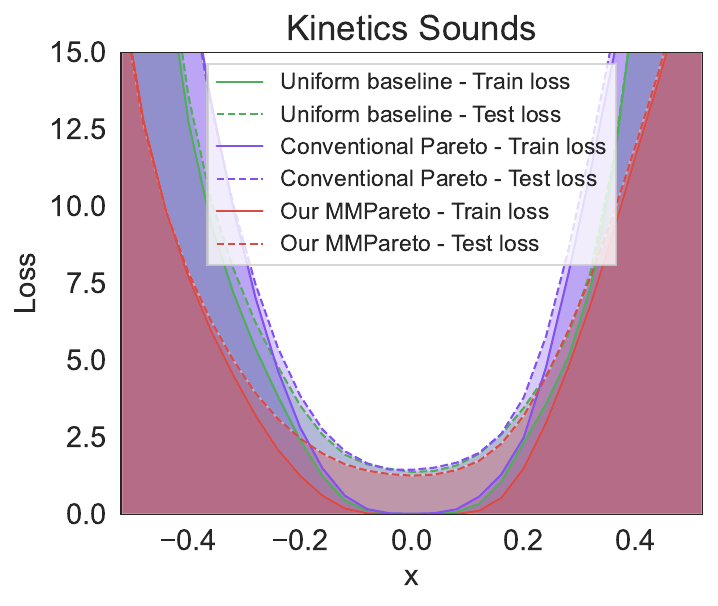}
			\caption{}
			\label{fig:ks-1d-loss}
	\end{subfigure}
	    \begin{subfigure}[t]{.23\textwidth}
			\centering
			\includegraphics[width=\textwidth]{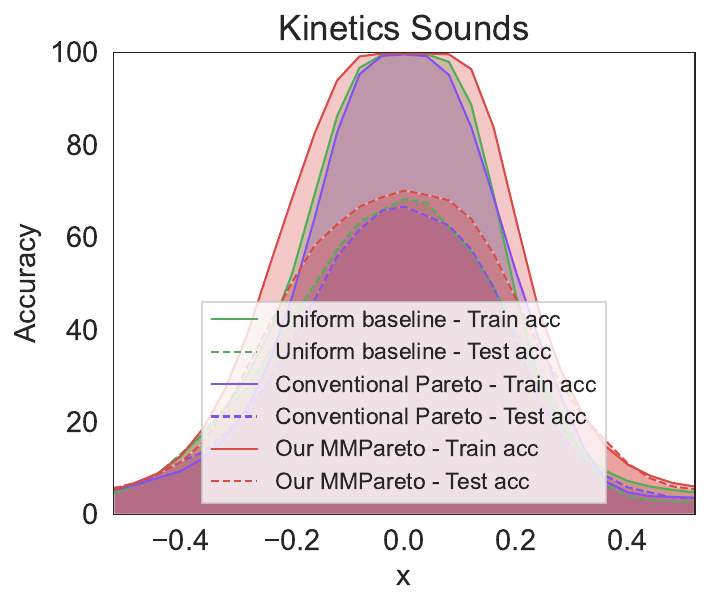}
			\caption{}
			\label{fig:ks-1d-acc}
	\end{subfigure}
 \vspace{-0.5em}
    \caption{Visualization of the loss landscape and corresponding accuracy of uniform baseline, conventional Pareto and our MMPareto methods. Our MMPareto method brings flatter minima. The visualization method is from~\citep{li2018visualizing}. Uniform baseline is a basic way where all losses are equally summed without special integration.}
    \vspace{-1em}
    \label{fig:losslandscpae}
\end{figure*}

\begin{table*}[t]
\centering
\caption{\textbf{Comparison with imbalanced multimodal learning methods where bold and underline represent the best and runner-up respectively.} * indicates that the unimodal evaluation (Acc audio and Acc video) is obtained by fine-tuning a unimodal classifier with frozen trained unimodal encoder, since this method could not provide unimodal prediction directly. This evaluation method borrows from~\cite{peng2022balanced}. $(\downarrow)$ indicates a performance drop compared with uniform baseline where all losses are equally summed.}
\label{tab:imbalance_cnn}
\setlength{\tabcolsep}{1.8mm}{
\begin{tabular}{c|ccc|ccc}
\bottomrule
\multirow{2}{*}{\textbf{Method}} & \multicolumn{3}{c|}{\textbf{CREMA-D}}                  & \multicolumn{3}{c}{\textbf{Kinetics Sounds}}           \\
                                 & \textbf{Acc} & \textbf{Acc audio} & \textbf{Acc video} & \textbf{Acc} & \textbf{Acc audio} & \textbf{Acc video} \\ \hline
Audio-only                       & -            & 61.69              & -                  & -            & \underline{53.63}              & -                  \\
Video-only                       & -            & -                  & \underline{56.05}              & -            & -                  & 49.20              \\
Unimodal pre-trained \& fine-tune                       & 71.51            & 60.08                 & \textbf{60.22}              &  68.75           &  53.49                & \underline{50.07}              \\
One joint loss*                   & 66.13        & 59.27              & 36.56              & 64.61        & 52.03              & 35.47              \\
Uniform baseline                 & 71.10         & \underline{63.44}              & 51.34              & 68.31        & 53.20              & 40.55              \\ \hline
G-Blending~\cite{wang2020makes}                       & \underline{72.01}        & 60.62 ($\downarrow$)              & 52.23              & \underline{68.90}        & 52.11 ($\downarrow$)             & 41.35              \\
OGM~\cite{peng2022balanced}*                              & 69.19 ($\downarrow$)        & 56.99 ($\downarrow$)              & 40.05 ($\downarrow$)              & 66.79 ($\downarrow$)        & 51.09 ($\downarrow$)              & 37.86 ($\downarrow$)              \\
Greedy~\cite{wu2022characterizing}*                           & 67.61 ($\downarrow$)       & 60.69 ($\downarrow$)            & 38.17 ($\downarrow$)              & 65.32 ($\downarrow$)        & 50.58 ($\downarrow$)              & 35.97 ($\downarrow$)             \\
PMR~\cite{fan2023pmr}*                              & 66.32 ($\downarrow$)        & 59.95 ($\downarrow$)              & 32.53 ($\downarrow$)              & 65.70 ($\downarrow$)        & 52.47 ($\downarrow$)              & 34.52 ($\downarrow$)              \\
AGM~\cite{li2023boosting}*                              & 70.06 ($\downarrow$)        & 60.38 ($\downarrow$)              & 37.54 ($\downarrow$)              & 66.17 ($\downarrow$)        & 51.31 ($\downarrow$)              & 34.83 ($\downarrow$)              \\ \hline
MMPareto                         & \textbf{75.13}        & \textbf{65.46}              & 55.24              & \textbf{70.13}        & \textbf{56.40}              & \textbf{53.05}              \\ \toprule
\end{tabular}}
\end{table*}

\subsection{Gradient Conflict and Magnitude Distribution in Multimodal Scenarios}

Here we verify the direction conflict and magnitude distribution between multimodal and unimodal gradient across different datasets. Firstly, in~\autoref{fig:teaser-conflic} and~\autoref{fig:cd-conflic-visual}, we show the cosine similarity between gradients on the Kinetics Sounds and CREMA-D. Based on the results, the update direction of multimodal and unimodal gradient indeed have conflict, \emph{i.e.,} negative cosine similarity, which potentially brings risk for the optimization of the corresponding shared unimodal encoder. In addition, such conflicts often exist in the early training stage, disturbance in which stage could substantially harm the model ability~\citep{liu2020early}. In addition, as~\autoref{fig:teaser-multimodal-distribution} and~\autoref{fig:cd-norm-visual}, we also observe the gradient magnitude distribution for a fixed set of parameters across different datasets. Each count is a different mini-batch. Based on the results, we could conclude that the multimodal gradient has a smaller magnitude and smaller batch sampling covariance, compared with the unimodal one. The difference in covariance could even be more than 3 times, greatly affecting model generalization.

\subsection{Loss Landscape Analysis}

Based on our analysis, conventional Pareto method could result in a sharper minima and then weakening model generalization. To empirically verify it, we visualize the loss landscape and the corresponding accuracy of uniform baseline, conventional Pareto method, and our MMPareto method. As shown in~\autoref{fig:losslandscpae}, conventional Pareto method indeed has a sharper minima, compared with uniform baseline where multimodal and unimodal losses are equally summed. Its loss and accuracy change more drastically. What's more, loss landscape near the minima that our MMPareto method converges on is the flattest one, since it ensures gradient with both non-conflict direction and enhanced SGD noise strength during optimization.

In addition, as shown in~\autoref{fig:losslandscpae}, the visually flatter minima consistently correspond to a lower test error. For example, our MMPareto method with flatter minima has a lower test error and higher accuracy. This phenomenon verifies that the flatter minima tend to generalize better, supporting our former analysis.

\begin{table*}[t]
\centering
\vspace{-0.5em}
\caption{\textbf{Comparison with imbalanced multimodal learning methods where bold and underline represent the best and runner-up respectively.} The network is transformer-based framework, MBT~\citep{nagrani2021attention}.}
\label{tab:imbalance_mbt}
\setlength{\tabcolsep}{2mm}{
\begin{tabular}{c|cccc|cccc}
\bottomrule
\multirow{3}{*}{\textbf{Method}} & \multicolumn{4}{c|}{\textbf{CREMA-D}}                                                   & \multicolumn{4}{c}{\textbf{Kinetics Sounds}}                                           \\
                                 & \multicolumn{2}{c}{\textbf{from scratch}} & \multicolumn{2}{c|}{\textbf{with pretrain}} & \multicolumn{2}{c}{\textbf{from scratch}} & \multicolumn{2}{c}{\textbf{with pretrain}} \\
                                 & \textbf{Acc}        & \textbf{macro F1}        & \textbf{Acc}         & \textbf{macro F1}         & \textbf{Acc}        & \textbf{macro F1}        & \textbf{Acc}         & \textbf{macro F1}        \\ \hline
One joint loss          &  44.96          &     42.78       &   66.69            &  67.26        &   42.51         &   41.56      & 68.30                & 69.31  \\
Uniform baseline                 & 45.30               & 43.74               & 69.89                & \underline{70.11}                & 43.31               & 43.08               & 69.40                & 69.60               \\ \hline
G-Blending~\cite{wang2020makes}                       & \underline{46.38}              &   \underline{45.16}            & \underline{69.91}                & 70.01                &   \underline{44.69}            &    \underline{44.19}            & 69.41                & 69.47               \\
OGM-GE~\cite{peng2022balanced}                           & 42.88               & 39.34               & 65.73                & 65.88                & 41.79               & 41.09               & 69.55                & 69.53               \\
Greedy~\cite{wu2022characterizing}                           & 44.49               & 42.76               & 66.67                & 67.26                & 43.31               & 43.08               & 69.62                & 69.75               \\
PMR~\cite{fan2023pmr}                              & 44.76               & 42.95               & 65.59                & 66.07                & 43.75               & 43.21               & \underline{69.67}                & \underline{69.87}               \\ 
AGM~\cite{li2023boosting}                              & 45.36               & 43.81               & 66.54                & 67.75                & 43.65               & 43.57               & 69.59                & 69.14               \\ \hline
MMPareto                         & \textbf{48.66}               & \textbf{48.17}               & \textbf{70.43}                & \textbf{71.17}                & \textbf{45.20}               & \textbf{45.26}               & \textbf{70.28}                & \textbf{70.11}                \\ \toprule
\end{tabular}}
\end{table*}

\begin{table*}[t]
\centering
\vspace{-1em}
\caption{\textbf{Comparison with related multi-task methods on Colored-and-gray-MNIST, ModelNet40 and Kinetics Sounds.} Bold and underline represent the best and runner-up respectively.}
\label{tab:grad}
\setlength{\tabcolsep}{2.5mm}{
\begin{tabular}{c|cc|cc|cc}
\bottomrule
\multirow{2}{*}{\textbf{Method}} & \multicolumn{2}{c|}{\textbf{\begin{tabular}[c]{@{}c@{}}CG-MNIST\end{tabular}}} & \multicolumn{2}{c|}{\textbf{\begin{tabular}[c]{@{}c@{}}ModelNet40\end{tabular}}} & \multicolumn{2}{c}{\textbf{\begin{tabular}[c]{@{}c@{}}Kinetics Sounds\end{tabular}}} \\
                                 & \textbf{Acc}                                 & \textbf{macro F1}                                 & \textbf{Acc}                                       & \textbf{macro F1}                                       & \textbf{Acc}                                     & \textbf{macro F1}                                        \\ \hline
One joint loss  &   60.50           &       59.89        &  87.88       &   83.32         & 64.61                     &     64.12            \\
Uniform baseline &     75.68         &  75.66    &  89.18 &  84.69     & 68.31                 &     68.13     \\ \hline
Conventional Pareto~\cite{sener2018multi}            &   62.00    &  61.85      &  88.05  &   83.01    &  66.64   & 66.17 \\
GradNorm~\cite{chen2018gradnorm}           &  76.16        &    76.12        & 88.98  &   83.79           &   65.84            &      65.14             \\
PCGrad~\cite{yu2020gradient}                &   \underline{79.35}               &   77.14             &  89.59  & 84.44         &       \underline{69.11}       &     \underline{68.75}             \\
MetaBalance~\cite{he2022metabalance}            &   79.18       &   \underline{77.87}       & \underline{89.63}     &    \underline{84.87}       &   68.90   & 68.62  \\  \hline
MMPareto              &  \textbf{81.88}              &     \textbf{81.69}         & \textbf{89.95} &   \textbf{85.15}         & \textbf{70.13}                &   \textbf{70.18}            \\  \toprule

\end{tabular}}
\end{table*}

\subsection{Comparison with Related Imbalanced Methods}

To validate the effectiveness of our MMPareto method in overcoming imbalanced multimodal learning problems, we compare it with recent studies: G-Blending~\citep{wang2020makes}, OGM-GE~\citep{peng2022balanced}, Greedy~\citep{wu2022characterizing}, PMR~\citep{fan2023pmr} and AGM~\citep{li2023boosting}. 

In addition, we also compare several basic settings. \textbf{Audio- and Video-only} are unimodal models trained individually. \textbf{Unimodal pre-trained \& fine-tune} is the method that fine-tunes the multimodal model with individually pre-trained unimodal encoders. \textbf{One joint loss} is the method that only uses multimodal joint loss. And \textbf{uniform baseline} is the method in which multimodal and unimodal losses are equally summed. To comprehensively evaluate the model ability, we further observe the unimodal performance, besides the common multimodal performance. 

Based on~\autoref{tab:imbalance_cnn}, we can find that the uniform baseline can achieve considerable performance, and even could outperform or be comparable with these imbalanced multimodal learning methods. The reason could be that the introduction of unimodal loss effectively enhances the learning of each modality, which accords with the core idea of these compared methods. Moreover, our MMpareto method with a conflict-free optimization process achieves a considerable improvement, compared with existing methods at the multimodal prediction. More than that, our MMPareto method simultaneously exhibits outstanding unimodal performance, and even can outperform individually trained unimodal model. For example, Audio accuracy of MMPareto is superior to Audio-only method on both CREMA-D and Kinetics Sounds datasets. This was rarely achieved in before studies.

Besides, we also conduct experiments under the widely used Transformer backbone, MBT~\citep{nagrani2021attention}, which contains both single-modal layers and cross-modal interaction layers. Compared to the former CNN backbone with the late fusion method, unimodal features in this transformer-based framework are more fully interacted and integrated. During experiments, we conduct experiments both from scratch and with ImageNet pre-training. Results are shown in~\autoref{tab:imbalance_mbt}. Based on the results, we can have the following observation. Firstly, former imbalanced multimodal learning could lose efficacy under these more complex scenarios with cross-modal interaction. For example, OGM-GE method is even worse than the one joint loss method on CREMA-D dataset. In contrast, our MMPareto gradient integration strategy is not only applicable to CNN backbones, but also able to maintain superior performance in transformer-based frameworks with complex interactions. In addition, whether or not to use pre-training does not affect the effectiveness of our method, which reflects its flexibility.

\begin{table*}[t]
\centering
 \vspace{-1em}
\caption{Comparison of multi-task methods on NYUv2 dataset. Conv Pareto is the conventional Pareto method.}
\label{tab:nyuv2}
\setlength{\tabcolsep}{1mm}{
\begin{tabular}{c|cccc}
\bottomrule
\multirow{2}{*}{\textbf{Method}} & \textbf{Segmentation - mIoU} & \textbf{Segmentation - Pix Acc} & \textbf{Depth - Abs Err} & \textbf{Depth - Rel Err} \\
                                 & (Higher Better)              & (Higher Better)                 & (Lower Better)           & (Lower Better)           \\ \hline
        Uniform baseline & 25.80 & 52.68 & 0.6309 & 0.2680 \\
        Conv Pareto~\cite{sener2018multi} & \textbf{27.66} & 53.91 & 0.6284 & 0.2685 \\ 
        GradNorm~\cite{chen2018gradnorm} & 26.17 & 53.41 & 0.6219 & 0.2738 \\
        PCGrad~\cite{yu2020gradient} & 26.44 & 53.93 & 0.6337 & 0.2658 \\ 
        MetaBalance~\cite{he2022metabalance} & 27.04 & \textbf{53.99} & 0.6258 & 0.2677 \\ \hline
        MMPareto & 26.35 & 53.48 & \textbf{0.6216} & \textbf{0.2656} \\ \toprule
    \end{tabular}}
\end{table*}

\begin{table}
\centering
 \vspace{-2em}
\caption{Results on MultiMNIST with $50\%$ salt-and-pepper noise on the right part of images. Uniform baseline is a basic way where all losses are equally summed. Conv Pareto is the conventional Pareto method.}
\label{tab:multitask}
\setlength{\tabcolsep}{0.6mm}{
\begin{tabular}{c|cc}
\bottomrule
\multirow{2}{*}{\textbf{Method}} & \multicolumn{2}{c}{\textbf{Accuracy}} \\
                        & \textbf{Task 1}        & \textbf{Task2}        \\ \hline
Uniform baseline        &  86.63        &   78.42      \\ \hline
Conv Pareto~\cite{sener2018multi}      &  86.95        &  77.04 ($\downarrow$)       \\
GradNorm~\cite{chen2018gradnorm} & 85.95($\downarrow$) & 79.52 \\
PCGrad~\cite{yu2020gradient} & 84.74($\downarrow$)  & 76.44($\downarrow$)  \\
MetaBalance~\cite{he2022metabalance} & 87.07 & 79.53 \\ \hline
MMPareto                &  \textbf{87.72}         &  \textbf{80.64}        \\ \toprule
\end{tabular}}
 \vspace{-1.5em}
\end{table} 

\subsection{Comparison with Related Multi-task Methods}

In past studies, there are other strategies that are used to balance multiple learning objectives. In this section, besides conventional Pareto~\cite{sener2018multi}, we also compare several representative ones: GradNorm~\citep{chen2018gradnorm}, PCGrad~\citep{yu2020gradient}, MetaBalance~\citep{he2022metabalance}. Experiments are conducted on different multimodal dataset, covering six types of modalities. Based on the results in~\autoref{tab:grad}, we further verify that the conventional Pareto method is inferior to the uniform baseline and loses its efficacy in multimodal scenarios. In addition, former multi-task methods are also possibly invalid in the context of multimodal learning. For example, GradNorm method is inferior to the uniform baseline on both ModelNet40 and Kinetics Sounds dataset. In contrast, our MMPareto method, which specifically considers the multimodal learning properties, maintains its superior performance across various dataset with different kinds of modalities.

\subsection{Extension to Multi-task Scenario}

To evaluate scalability of our method in multi-task cases with similar property that there is a clear discrepancy in task difficulty, we conduct experiments on MultiMNIST dataset~\citep{sabour2017dynamic}. In MultiMNIST, two images with different digits from the original MNIST dataset are picked and then combined into a new one by putting one digit on the left and the other one on the right. Two tasks are to classify these two digits. To increase the difference in difficulty of tasks, we add $50\%$ salt-and-pepper noise on the right part of images. Several data samples are provided in~\autoref{sec:smaple_mm}. Based on~\autoref{tab:multitask}, most multi-task methods have a performance drop (\emph{e.g.,} GradNorm, PCGrad and conventional Pareto). Not surprisingly, our MMPareto could extend to this scenario and achieve considerable performance, indicating its ideal scalability.

Besides the MultiMNIST dataset, we also conduct experiments on the typical multi-task dataset, NYUv2. Widely-used benchmark~\citep{liu2019end} is used. We consider two tasks, semantic segmentation and depth estimation. As shown in~\autoref{tab:nyuv2}, although our method is built on multimodal properties, it also achieves improvement on typical multi-task dataset, compared with uniform baseline. It even achieves the best results on depth estimation task. These results show the versatility of our method.

\section{Discussion}

In this paper, we identify previously ignored gradient conflicts in multimodal scenarios with discrepancies in learning difficulty of uni- and multimodal objectives, then propose MMPareto algorithm to diminish these conflict and alleviate imbalanced multimodal learning. Besides typical discriminative multimodal scenarios, more multimodal cases, like multimodal pre-training, have witnessed rapid development recently~\cite{wang2023large,zhang2023multimodal,chen2023vlp}. Our method is also expected to apply to these more general multimodal scenarios, like BLIP-2, easing the potential conflicts among multiple learning objectives. 

\textbf{Future work and limitation:} We strengthen SGD noise term to improve model generalization, by adjusting gradient magnitude in this paper. In fact, the gradient covariance $\frac{1}{|S|} C$ of multiple losses and their correlation are also expected to provide a reliable reference when ensuring model generalization during gradient integration. Hence how to efficiently estimate and utilize these covariance is a promising next direction. Besides, our current convergence analysis is with relatively ideal assumptions. But in practice, it can greatly improve multimodal learning with a modest number of iterations in an acceptable range. This also inspires us to explore multimodal Pareto methods with more rigorous convergence theory.

\section*{Acknowledgements}
This research was supported by National Natural Science Foundation of China (NO.62106272), and Public Computing Cloud, Renmin University of China.

\section*{Impact Statement}
Our method could also contribute to our society in several aspects, \emph{e.g.,} helping autonomous vehicle by enhancing multi-sensory learning.

\nocite{langley00}

\bibliography{example_paper}
\bibliographystyle{icml2024}

\clearpage
\appendix

\section{Parameter Description}
For convenience, we include a table of parameter descriptions in~\autoref{tab:parameter}.

\section{Dataset and Experiment Settings}
\label{sec:dataset}

\noindent \textbf{CREMA-D}~\citep{cao2014crema} is an audio-visual dataset for emotion recognition, including 7,442 video clips, each spanning 2 to 3 seconds in duration. The video content is that actors speak several short words. This dataset covers 6 emotions: angry, happy, sad, neutral, discarding, disgust and fear. 

\noindent \textbf{Kinetics Sounds}~\citep{arandjelovic2017look} is an audio-visual dataset containing 31 human action classes selected from Kinetics dataset~\citep{kay2017kinetics}. All videos are manually annotated for human action using Mechanical Turk and cropped to 10 seconds long around the action. 

\noindent \textbf{Colored-and-gray-MNIST}~\citep{kim2019learning} is a synthetic dataset based on MNIST~\citep{lecun1998gradient}. Each instance contains two kinds of images, a gray-scale and a monochromatic colored image. Monochromatic images in the training set are strongly color-correlated with their digit labels, while monochromatic images in the other sets are weakly color-correlated with their labels.

\noindent \textbf{ModelNet40}~\citep{wu20153d} is a dataset with 3D objects, covering 40 categories. It contains 9,483 training samples and 2,468 test samples. This dataset could be used to classify these 3D objects based on the 2D views of their front-view and back-view data~\citep{su2015multi}. Data of all views is a collection of 2D images of a 3D object.

When not specified, ResNet-18~\citep{he2016deep} is used as the backbone in experiments and models are trained from scratch. Concretely, for the visual encoder, we take multiple frames as the input, and feed them into the 2D network; for the audio encoder, we modified the input channel of ResNet-18 from three to one like~\cite{chen2020vggsound} does and the rest parts remain unchanged; Encoders of other modalities are not modified. For the CNN backbone, we use the widely used late fusion method, to integrate unimodal features. For the Transformer backbone, MBT~\citep{nagrani2021attention}, is used as the backbone. Concretely, the backbone contains 6 single-modal layers and 2 layers with cross-modal interaction. Specifically, for the Colored-and-gray MNIST dataset, we build a neural network with 4 convolution layers and 1 average pool layer as the encoder, like~\cite{fan2023pmr} does.  During the training, we use SGD with momentum ($0.9$) and set the learning rate at $1e-3$. All models are trained on 2 NVIDIA RTX 3090 (Ti).

\section{Samples of MultiMNIST Dataset}
\label{sec:smaple_mm}
Here we provide several samples of MultiMNIST dataset. In MultiMNIST, two images with different digits from the original MNIST dataset are picked, and then combined into a new one by putting one digit on the left and the other one on the right. Two tasks are to classify these two digits. In order to increase the difference in difficulty between tasks, we add $50\%$ salt-and-pepper noise on the right part of images.

\begin{figure}[h]
\centering
            \begin{subfigure}[t]{.3\textwidth}
			\centering
			\includegraphics[width=0.3\textwidth]{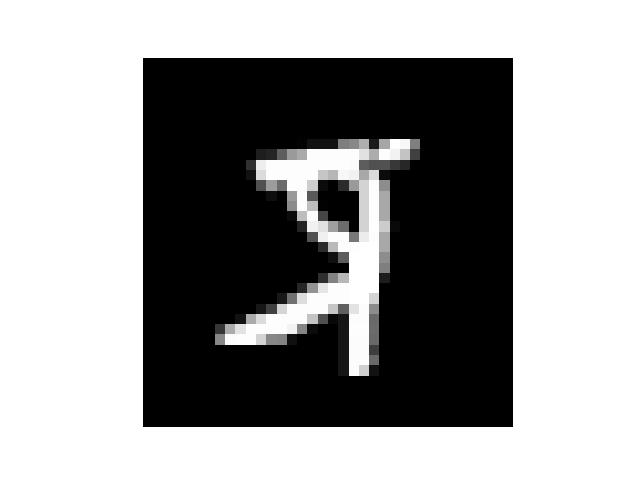}
			\caption{Sample 1 without noise; Label: [5,1].}
	\end{subfigure}
	    \begin{subfigure}[t]{.3\textwidth}
			\centering
			\includegraphics[width=0.3\textwidth]{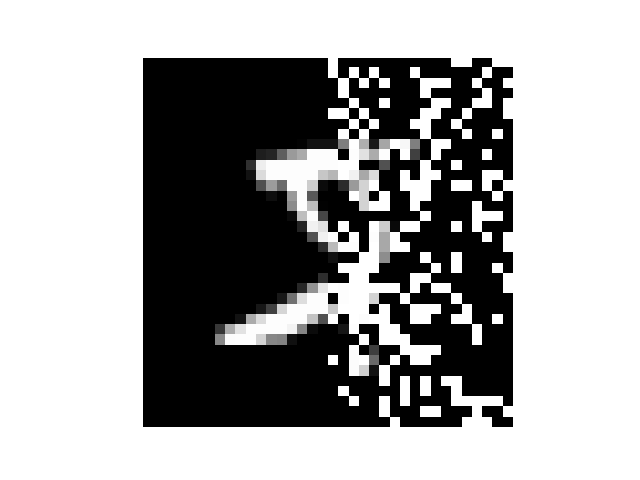}
			\caption{Sample 1 with noise; Label: [5,1].}
	\end{subfigure}

            \begin{subfigure}[t]{.3\textwidth}
			\centering
			\includegraphics[width=0.3\textwidth]{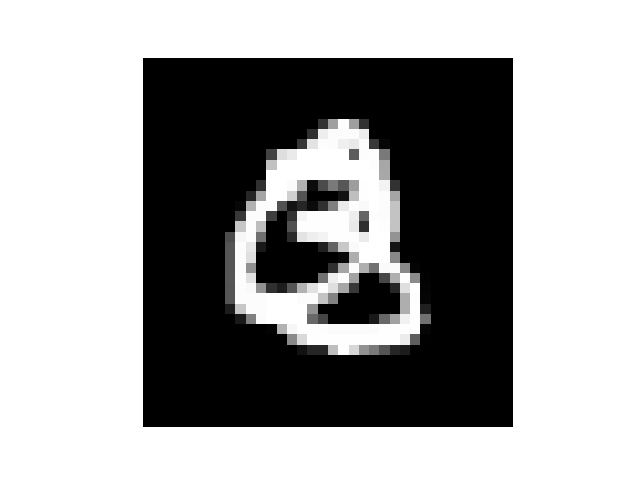}
			\caption{Sample 2 without noise; Label: [0,3].}
	\end{subfigure}
	    \begin{subfigure}[t]{.3\textwidth}
			\centering
			\includegraphics[width=0.3\textwidth]{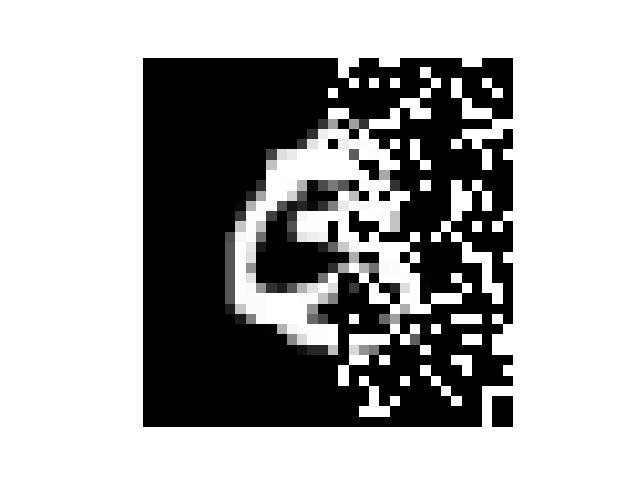}
			\caption{Sample 2 with noise; Label: [0,3].}
	\end{subfigure}
    \caption{Samples in MultiMNIST dataset.}
   
\end{figure}

\begin{table*}[t]
\centering
\caption{The description of used parameters.}
\label{tab:parameter}
\begin{tabular}{c|c}
\bottomrule
Parameter & Description \\ \hline
$n$                 & total number of modalities      \\
${\mathcal{L}}_m$   & multimodal joint loss       \\
${\mathcal{L}}_u^k$ & unimodal loss of $k-$th modality    \\
$N$                 & number of training samples  \\
$X_i$        & input of $i-$th data sample            \\
$Y_i$        & label of $i-$th data sample            \\
$(X_i, Y_i)$        & $i-$th data sample            \\
$\theta^k(t)$       & parameters of unimodal encoder of the $k-$th modality at $t-$th iteration \\
$S$ & set of $t-$th mini-batch\\
$\frac{1}{\|S\|}C^m$   &  batch sampling covariance of multimodal joint loss \\
$\frac{1}{\|S\|} C^u$ & batch sampling covariance of unimodal loss \\
$\mathbf{g}^m_S$ / $\mathbf{g}^m_S(\theta^k(t))$& multimodal gradient of $t-$th iteration for encoder of $k-$th modality  \\
$\mathbf{g}^u_S$ / $\mathbf{g}^u_S(\theta^k(t))$& unimodal gradient of $t-$th iteration for encoder of $k-$th modality  \\
$\alpha^m$ & weight of multimodal gradient $\mathbf{g}^m_S$ in Pareto optimization problem \\
$\alpha^u$ & weight of unimodal gradient $\mathbf{g}^u_S$ in Pareto optimization problem \\
$\beta$ & the angle between $\mathbf{g}^m_S$ and $\mathbf{g}^u_S$ \\
$\eta$  & learning rate of SGD optimization \\
$\gamma$  & rescale factor of the integrated gradient of MMPareto method  \\
$\mathbf{h}_S(\theta^k(t))$ & the final integrated gradient of uniform baseline at $t-$th iteration\\
$\epsilon_t$ & SGD noise term of uniform baseline at $t-$th iteration \\
$\mathbf{h}^{\text{Pareto}}_S(\theta^k(t))$ & the final integrated gradient of conventional Pareto method at $t-$th iteration \\
$\zeta_t$ & SGD noise term of conventional Pareto method at $t-$th iteration \\
$\mathbf{h}^{\text{MMPareto}}_S(\theta^k(t))$ & the final integrated gradient of proposed MMPareto method at $t-$th iteration \\
$\xi_t$ & SGD noise term of proposed MMPareto method at $t-$th iteration \\ \toprule
    \end{tabular}
\end{table*}

\section{More-than-two Modality Case}
\begin{table}[h]
\centering
\caption{Comparison of imbalanced multimodal learning methods on CMU-MOSI dataset with three modalities. Greedy method could not extend to more-than-two modality case. * indicates that the original methods of OGM and PMR also only consider two modality cases, and we extend them while retaining core strategy.}
\label{tab:three}
\begin{tabular}{c|cc}
\toprule
\textbf{Method}  & \textbf{Acc}   & \textbf{macro F1} \\ \midrule[0.7pt]
One joint loss   & 75.07          & 73.16             \\
Uniform baseline & 75.95          & 73.93             \\ \midrule
G-Blending~\cite{wang2020makes}       & 76.16          & 74.65             \\
OGM*~\cite{peng2022balanced}             & 75.80          & 74.71             \\
Greedy~\cite{wu2022characterizing}           & /              & /                 \\
PMR*~\cite{fan2023pmr}             & 76.28          & 75.06             \\
AGM~\cite{li2023boosting}              & 76.08          & 74.98             \\ \midrule
MMPareto         & \textbf{76.53} & \textbf{75.59}    \\ \bottomrule
\end{tabular}
\end{table}
Our MMPareto method has no restriction on the number of modalities. To verify the method performance in more-than-two modality case, we conduct experiments on CMU-MOSI dataset~\cite{zadeh2016mosi}. CMU-MOSI is a sentiment analysis dataset with three modalities, audio, video and text. It is annotated with utterance-level sentiment labels. This dataset consists of 93 movie review videos segmented into 2,199 utterances. According to results in~\autoref{tab:three}, many existing methods for imbalanced multimodal learning problem only consider two modality case, and even can not extend to more modalities cases. Our MMPareto method is free of the limitation of the number of modalities and maintains effectiveness.

\section{Proof for the Convergence of MMPareto}
\label{sec:convergence}
\setcounter{remark}{1}
\begin{remark}
The proposed MMPareto method admits an iteration sequence that converges to a Pareto stationarity.
\end{remark}
\newpage

\begin{proof}
In MMPareto algorithm, at each training iteration, we first sovle the optimization problem:
\begin{equation}
\label{equ:proof_pareto}
\begin{gathered}
\min_{\alpha^m,\alpha^u \in \gR} {\| \alpha^m \rvg^m_S +\alpha^u \rvg^u_S \|}^2 \\
s.t. \quad \alpha^m,\alpha^u \geq 0, \alpha^m+\alpha^u=1.
\end{gathered}
\end{equation}
For brevity, here we use $\{\rvg^i_S\}_{i \in \{m,u\}}$ to substitute $\{\rvg^i_S(\theta^k(t))\}_{i \in \{m,u\}}$. $\| \cdot \|$ denotes the $L_2$-norm. This problem is equal to finding the minimum-norm in the convex hull of the family of gradient vectors $\{\rvg^i_S\}_{i \in \{m,u\}}$. We denote the found minimum-norm as $\omega=\alpha^m \rvg^m_S +\alpha^u \rvg^u_S$. Based on~\cite{desideri2012multiple}, either $\omega$ to this optimization problem is $0$ and the corresponding parameters are Pareto-stationary which is a necessary condition for Pareto-optimality, or $\omega$ can provide a descent directions common to all learning objectives. When the minimum-norm $\omega$ does not satisfy the condition of Pareto stationarity, we consider the non-conflict case and conflict case respectively. 

We first analyze the non-conflict case, where $\cos\beta \geq 0$. $\beta$ is the angle between $\rvg^m_S$ and $\rvg^u_S$. Under this case, we assign $2\alpha^m = 2\alpha^u = 1$. Then, the final gradient $\rvh^{\text{Pareto}}_S=\rvg^m_S + \rvg^u_S$ is with direction that can benefit all losses and enhanced generalization. Then we analyze the conflict case, where $\cos\beta < 0$. The results of optimization problem~\autoref{equ:proof_pareto} are used as $\alpha^m$ and $\alpha^u$. Based on the above statement, we can have that $\rvh^{\text{Pareto}}_S= 2\alpha^m \rvg^m_S +2\alpha^u \rvg^u_S$ can provide a direction that is common to all learning objective. Furthermore, we enhance the magnitude of final gradient to enhance SGD noise strength for improving model generalization. \emph{In summary, the final gradient of MMPareto could always provide the direction that is common to all learning objectives.} 

If the training iteration stops in a finite number of steps, a Pareto-stationary point has been reached. Otherwise, the iteration continues indefinitely, generating an infinite sequence of shared parameters ${\theta^k}$. \emph{Since the value of loss function $\gL_m$ and $\gL_{u}^k$ is positive and monotone-decreasing during optimization, it is bounded. Hence, the sequence of parameter ${\theta^k}$ is itself bounded and it admits a subsequence converging to ${\theta^k}^*$.}

Necessarily, ${\theta^k}^*$ is a Pareto-stationary point. In other words, the minimum-norm $\omega^*$ is zero at this step. To establish this, assume instead that the obtained minimum-norm $\omega^*$, which corresponds to ${\theta^k}^*$, is nonzero. A new iteration would potentially diminish each learning objective of a finite amount, and a better solution of parameter $\theta^{k}$ be found. 

Overall, the proposed MMPareto method admits an iteration sequence that converges to a Pareto stationarity.

\end{proof}

\section{Computation Cost and Convergence of MMPareto}
\label{sec:converge-exp}
In this section, we observe the time cost of gradient weight computation process and the convergence of our MMPareto method. In~\autoref{tab:cost}, we record the mean and variance of time cost per iteration. Based on the results, this process would not take a lot of time, without much effect on total training time.

In addition, we observe the number of iterations it takes to find the Pareto stationarity. According to results in~\autoref{tab:convergence}, our method typically converges in a modest number of iterations in an acceptable range.

\begin{table}[h]
\caption{Time cost of MMPareto calculation per iteration.}
\label{tab:cost}
\setlength{\tabcolsep}{1mm}{
\begin{tabular}{c|c|c}
\toprule
\textbf{Dataset} & \textbf{Time cost mean} & \textbf{Time cost variance} \\ \midrule[0.7pt]
CREMA-D          & 0.11s             & 1e-5                        \\
Kinetics Sounds  & 0.16s            & 1e-5                        \\
CG-MNIST         & 0.05s             & 1e-5                        \\
ModelNet40       & 0.10s             & 1e-5                        \\ \bottomrule
\end{tabular}}
\end{table}
 
\begin{table}[h]
\caption{Number of iterations that find Pareto stationarity.}
\label{tab:convergence}
\begin{tabular}{c|c|c}
\toprule
\textbf{Dataset} & \textbf{Audio encoder} & \textbf{Video encoder} \\ \midrule[0.7pt]
CREMA-D          & \#3098                 & \#1709                 \\
Kinetics Sounds  &  \#10197                  &  \#7957                      \\ \bottomrule
\end{tabular}
\end{table}


\end{document}